\documentclass[pdflatex,sn-mathphys-num]{sn-jnl}

\usepackage{graphicx}%
\usepackage{multirow}%
\usepackage{amsmath,amssymb,amsfonts}%
\usepackage{amsthm}%
\usepackage{mathrsfs}%
\usepackage[title]{appendix}%
\usepackage{xcolor}%
\usepackage{textcomp}%
\usepackage{manyfoot}%
\usepackage{booktabs}%
\usepackage{algorithm}%
\usepackage{algorithmicx}%
\usepackage{algpseudocode}%
\usepackage{listings}%

\theoremstyle{thmstyleone}%
\theoremstyle{thmstyletwo}%

\theoremstyle{thmstylethree}%

\begin{document}

\title[Beyond Accuracy: Evaluating Forecasting Models by Multi-Echelon Inventory Cost]{Beyond Accuracy: Evaluating Forecasting Models by Multi-Echelon Inventory Cost}


\author*[1]{\fnm{Swata} \sur{Marik}}\email{swatamarik01@gmail.com}
\equalcont{These authors contributed equally to this work.}

\author[2]{\fnm{Swayamjit} \sur{Saha}}\email{ss4706@msstate.edu}
\equalcont{These authors contributed equally to this work.}

\author[3]{\fnm{Garga} \sur{Chatterjee}}\email{garga@isical.ac.in}

\affil*[1]{\orgdiv{Department of Home Science}, \orgname{University of Calcutta}, \orgaddress{\street{20 E, Judges Court Road, Alipore}, \city{Kolkata}, \postcode{700027}, \state{West Bengal}, \country{India}}}

\affil[2]{\orgdiv{Department of Computer Science and Engineering}, \orgname{Mississippi State University}, \orgaddress{\street{665 George Perry Street}, \city{Starkville}, \postcode{39759}, \state{MS}, \country{USA}}}

\affil[3]{\orgdiv{Psychology Research Unit}, \orgname{Indian Statistical Institute Kolkata}, \orgaddress{\street{203, B.T. Road}, \city{Kolkata}, \postcode{700108}, \state{West Bengal}, \country{India}}}


\abstract{This study develops a digitalized forecasting--inventory optimization pipeline integrating traditional forecasting models, machine learning regressors, and deep sequence models within a unified inventory simulation framework. Using the M5 Walmart dataset, we evaluate seven forecasting approaches and assess their operational impact under single- and two-echelon newsvendor systems. Results indicate that Temporal CNN and LSTM models significantly reduce inventory costs and improve fill rates compared to statistical baselines. Sensitivity and multi-echelon analyses demonstrate robustness and scalability, offering a data-driven decision-support tool for modern supply chains.}




\maketitle

\section{Introduction}\label{sec1}

Modern supply chains face growing volatility and disruption risk, motivating digitalization and AI-enabled decision support to improve resilience. Digital transformation allows firms to integrate heterogeneous data, automate analytics, and respond adaptively to shocks; recent work links ML-driven digitalization to stronger disruption absorption and service continuity \cite{ivanov2020viability, queiroz2020impacts}. Within this ecosystem, demand forecasting is a key input to inventory optimization, shaping replenishment, safety stock, and coordination across distribution centers (DCs) and stores. Forecast errors can propagate upstream and amplify variability (the bullwhip effect) \cite{lee1997bullwhip}, so improved forecasts can directly reduce cost and improve service \cite{hyndman2018forecasting}.

However, classical single-model approaches (e.g., ARIMA and exponential smoothing) often struggle with nonlinear, intermittent, and non-stationary retail demand \cite{makridakis2018statistical}. ML/DL models (e.g., recurrent and attention-based architectures) frequently outperform statistical baselines under such conditions \cite{bandara2020rnnclusters, lim2021tft, lim2021dl_survey}, yet their value is commonly reported via error metrics rather than operational impact \cite{syntetos2009review, petropoulos2019m3inventory, jeunet2006multilevel}. This gap is more pronounced in multi-echelon networks: despite the importance of two-echelon DC--store systems \cite{nahmias1994twoechelon}, relatively few empirical studies link forecast improvements to coordinated multi-tier inventory outcomes using real retail data and information-sharing signals \cite{vanbelle2021sellthrough, schlaich2024nextorder, abolghasemi2025posorders}.

To address this gap, this study makes four key contributions. First, we develop a unified forecasting pipeline that integrates seven model classes—including statistical, machine learning, and deep learning approaches—within a standardized feature and training framework. Second, we embed these forecasts into a newsvendor‑based operational evaluation to quantify how predictive performance translates into practical inventory outcomes across multiple holding–shortage cost ratios. Third, we extend the analysis beyond traditional single-echelon settings by implementing a two‑echelon DC–Store simulation, enabling assessment of upstream and downstream operational impacts. Finally, we perform a detailed sensitivity analysis to evaluate the robustness of each forecasting model under varying cost structures, offering actionable insights for practitioners seeking to leverage digital analytics for supply chain resilience.

\section{Literature Review}

\subsection{Classical Forecasting}
Classical forecasting is widely used in supply chains due to simplicity and ease of deployment. Common baselines include ARIMA and exponential smoothing (e.g., Holt--Winters) \cite{hyndman2008forecast, chatfield1988holtwinters, gardner2006es2}. However, retail demand often violates linearity/stationarity assumptions, especially under intermittency and structural breaks. Croston-type methods address intermittent demand by separating demand sizes and inter-arrival times \cite{croston1972intermittent, syntetos2005intermittent}, yet can still degrade under abrupt shifts \cite{syntetos2009review}.

\subsection{Machine Learning Approaches}
Tree ensembles such as gradient boosting and related methods are popular for retail forecasting because they capture nonlinear interactions among engineered features and calendar signals \cite{friedman2002sgb}. Scalable implementations like XGBoost have shown strong performance on large tabular demand datasets \cite{chen2016xgboost}, and evidence from the M5 competition highlights the competitiveness of boosted-tree approaches for hierarchical retail sales prediction \cite{makridakis2022m5back, makridakis2022m5acc}.

\subsection{Deep Learning Approaches}
Deep learning models learn temporal representations directly from sequences and covariates, which can reduce reliance on handcrafted features. LSTMs are a standard choice for long-range dependencies \cite{hochreiter1997lstm}, while multi-horizon architectures such as Temporal Fusion Transformers improve covariate handling and interpretability \cite{lim2021tft}. Temporal CNNs with dilations provide efficient long-context modeling \cite{borovykh2017cnn}, and hybrid ARIMA--NN models have been used to combine linear structure with nonlinear effects \cite{zhang2003hybrid}. Surveys emphasize DL advantages when demand drivers interact and forecasting must scale across many related series \cite{lim2021dl_survey, bandara2020rnnclusters}.

\subsection{Gaps in Prior Work}

\subsubsection{Forecasting accuracy vs.\ inventory outcomes} Forecasting studies commonly emphasize statistical error measures, while inventory studies often assume exogenous/stylized demand inputs; consequently, the operational value of accuracy improvements is not consistently quantified in cost and service metrics \cite{syntetos2009review, petropoulos2019m3inventory}.
\subsubsection{Limited multi-echelon evaluation with real retail data} Although multi-echelon theory is well developed, empirical studies that propagate forecast errors through DC--store replenishment decisions using real retail demand and information-sharing signals (e.g., POS/sell-through) remain comparatively scarce \cite{nahmias1994twoechelon, vanbelle2021sellthrough, schlaich2024nextorder}.
\subsubsection{Lack of unified comparisons across statistical, ML, and DL} Only a small subset of work evaluates classical methods, ML ensembles, and DL architectures under a \emph{single} experimental protocol that links accuracy to downstream multi-tier inventory performance (e.g., total cost, fill rate, backorders) \cite{petropoulos2019m3inventory, jeunet2006multilevel, abolghasemi2025posorders}.

\section{Methodology}

\subsection{Dataset and Preprocessing}
We use the M5 Forecasting dataset \cite{kaggle_m5_forecasting_accuracy}, combining daily sales from \texttt{sales\_train\_validation.csv} with exogenous covariates from \texttt{calendar.csv}. Sales are reshaped to long format and merged with the calendar on the \texttt{d} index (date, weekday/month, events, SNAP).

\begin{itemize}
\item \textbf{Subset:} We filter to \texttt{state\_id=CA} and \texttt{dept\_id=FOODS\_1} (CA\_FOODS\_1) for controlled benchmarking.
\item \textbf{Features:} Standard retail predictors are constructed per series: lags ($y_{t-1},y_{t-7},y_{t-14},y_{t-28}$), rolling means (7/14/28 days), and calendar/event indicators (including SNAP).
\item \textbf{Splits:} We apply a rolling holdout: final 28 days for test, previous 28 days for validation, remainder for training, with all features computed from past data only.
\end{itemize}

\subsection{Forecasting Models}
Let $y_{i,t}$ be daily demand for series $i$ at time $t$, and $\hat{y}_{i,t+1|t}$ the one-step-ahead forecast. Classical models are fit per series, while ML/DL models are trained globally across the selected panel using engineered covariates $x_t$ (lags, rolling statistics, and calendar/event indicators). All models are trained on the training split, tuned on validation, and evaluated on test.

\begin{enumerate}
\item \textbf{Naive (lag-1):} a persistence baseline that is often competitive for very short horizons.
\begin{equation}
\hat{y}_{i,t+1|t}=y_{i,t}.
\end{equation}

\item \textbf{Holt--Winters ES (additive, $m=7$):} captures level, trend, and weekly seasonality via exponential smoothing.
\begin{align}
\ell_t &= \alpha(y_t-s_{t-m})+(1-\alpha)(\ell_{t-1}+b_{t-1}),\\
b_t &= \beta(\ell_t-\ell_{t-1})+(1-\beta)b_{t-1},\\
s_t &= \gamma(y_t-\ell_t)+(1-\gamma)s_{t-m},\\
\hat{y}_{t+1|t} &= \ell_t+b_t+s_{t+1-m}.
\end{align}

\item \textbf{ARIMA(1,1,1):} a differenced linear time-series model that captures short-range autocorrelation in the mean.
\begin{equation}
\nabla y_t = c + \phi_1\nabla y_{t-1} + \varepsilon_t + \theta_1\varepsilon_{t-1}.
\end{equation}

\item \textbf{Gradient Boosting Regressor (GBR):} an ensemble of regression trees that learns nonlinear relations between $x_t$ and demand.
\begin{equation}
\hat{y}_{t+1|t}=f(x_t),\qquad f(x)=\sum_{m=1}^{M}\eta\,g_m(x).
\end{equation}

\item \textbf{XGBoost:} a regularized gradient-boosted tree model optimized for scalability and strong tabular performance.
\begin{equation}
\hat{y}_{t+1|t}=\sum_{k=1}^{K} f_k(x_t),\qquad 
\mathcal{L}=\sum_t \ell(y_t,\hat{y}_t)+\sum_k \Omega(f_k).
\end{equation}

\item \textbf{LSTM (global):} a recurrent sequence model that learns temporal representations shared across series.
\begin{align}
c_t &= f_t\odot c_{t-1}+i_t\odot \tilde{c}_t,\qquad
h_t=o_t\odot\tanh(c_t),\\
\hat{y}_{t+1|t} &= W_y h_t + b_y.
\end{align}

\item \textbf{Temporal CNN:} a causal dilated-convolution model with large receptive fields to capture long-range patterns efficiently.
\begin{align}
z_t^{(\ell)} &= \sum_{k=0}^{K-1} w_k^{(\ell)}\,x_{t-d_\ell k}^{(\ell-1)},\\
\hat{y}_{t+1|t} &= W_z z_t^{(L_c)} + b_z.
\end{align}
\end{enumerate}

\subsection{Newsvendor Inventory Simulator}
We evaluate each forecasting model through a rolling single-period \emph{newsvendor} simulator. Let $D_{i,t}$ be realized demand for series $i$ at day $t$, and let $Q_{i,t}$ be the order placed before observing $D_{i,t}$. With overage (holding) cost $h>0$ and underage (shortage) cost $b>0$, the per-period cost is
\begin{equation}
C_{i,t}(Q_{i,t})=h\max(Q_{i,t}-D_{i,t},0)+b\max(D_{i,t}-Q_{i,t},0).
\end{equation}

\begin{itemize}
\item \textbf{Mapping forecasts to orders:} for point forecasts $\hat{D}_{i,t|t-1}$, we set
\begin{equation}
Q_{i,t}=\max\{0,\hat{D}_{i,t|t-1}\}.
\end{equation}

\item \textbf{KPIs:} we report mean cost over the evaluation horizon $\mathcal{T}$,
\begin{equation}
\overline{C}=\frac{1}{N|\mathcal{T}|}\sum_{i=1}^{N}\sum_{t\in\mathcal{T}} C_{i,t}(Q_{i,t}),
\end{equation}
and demand-weighted fill rate,
\begin{equation}
\mathrm{FR}=1-\frac{\sum_{i,t\in\mathcal{T}} \max(D_{i,t}-Q_{i,t},0)}{\sum_{i,t\in\mathcal{T}} D_{i,t}+\epsilon}.
\end{equation}
\end{itemize}

\subsection{Two-Echelon Extension}
We extend the simulator to a two-echelon system with one DC supplying stores $\mathcal{S}$. DC demand is aggregated as
\begin{equation}
D^{DC}_t=\sum_{s\in\mathcal{S}} D_{s,t}, \qquad 
\hat{D}^{DC}_{t|t-1}=\sum_{s\in\mathcal{S}} \hat{D}_{s,t|t-1},
\end{equation}
and the DC orders $Q^{DC}_t=\max\{0,\hat{D}^{DC}_{t|t-1}\}$. Stores request $R_{s,t}=\max\{0,\hat{D}_{s,t|t-1}\}$; if inventory is insufficient, fulfillment is allocated proportionally:
\begin{equation}
F_{s,t}= \frac{R_{s,t}}{\sum_{s'}R_{s',t}+\epsilon}\,\big(I^{DC}_t+Q^{DC}_t\big).
\end{equation}
Two-echelon performance is measured by average network cost $\overline{C}^{(2)}$ and fill rate:
\begin{equation}
\overline{C}^{(2)}=\frac{1}{|\mathcal{T}|}\sum_{t\in\mathcal{T}}\Big(C^{DC}_t+\sum_{s\in\mathcal{S}} b^{S}\max(D_{s,t}-F_{s,t},0)\Big),
\end{equation}
\begin{equation}
\mathrm{FR}^{(2)}=
\frac{\sum_{t\in\mathcal{T}}\sum_{s\in\mathcal{S}} \min(D_{s,t},F_{s,t})}
{\sum_{t\in\mathcal{T}}\sum_{s\in\mathcal{S}} D_{s,t}+\epsilon}.
\end{equation}

\section{Results}
We report results on the held-out \textbf{test} period for the CA\_FOODS\_1 subset, evaluating both (i) statistical forecast accuracy and (ii) downstream inventory performance under the newsvendor simulator with overage cost $h=1$ and underage cost $b\in\{2,5,10\}$.

\subsection{Forecast accuracy and single-echelon inventory performance}
Table~\ref{tab:results_main} summarizes test-set accuracy (RMSE/MAE) together with inventory KPIs (average daily newsvendor cost and fill rate) under $(h,b)=(1,5)$. Overall, modern learning-based models yield consistent operational gains over classical baselines. The \textbf{Temporal CNN} achieves the \emph{lowest} inventory cost ($3.674$) and the \emph{highest} fill rate ($0.632$), corresponding to an \textbf{18.7\%} cost reduction and a \textbf{+9.8 pp} fill-rate improvement over the naive forecast. The \textbf{LSTM} attains the best RMSE ($2.207$) and delivers the second-best inventory cost ($3.704$). Tree ensembles (GBR/XGBoost) also improve both cost and service compared to classical and naive baselines. We note that MAPE is extremely large for this subset due to near-zero denominators on low-demand days; hence we emphasize RMSE/MAE and inventory-based metrics for interpretation.

\begin{table}[t]
\centering
\small
\caption{Test-set forecast accuracy and inventory performance (single-echelon newsvendor, $h=1$, $b=5$). Cost reduction and fill-rate gains are relative to Naive (lag-1).}
\label{tab:results_main}
\begin{tabular}{lcccccc}
\toprule
Model & RMSE & MAE & Avg. Cost/day & Fill rate & Cost $\downarrow$ & Fill $\uparrow$ (pp) \\
\midrule
Naive (lag-1)         & 2.909 & 1.505 & 4.521 & 0.534 & 0.0\%  & 0.0 \\
Holt--Winters ES      & 2.677 & 1.487 & 4.182 & 0.583 & 7.5\%  & 4.9 \\
ARIMA(1,1,1)          & 2.636 & 1.486 & 4.258 & 0.572 & 5.8\%  & 3.8 \\
GBR (global)          & 2.296 & 1.293 & 3.854 & 0.605 & 14.8\% & 7.1 \\
XGBoost (global)      & 2.294 & 1.289 & 3.839 & 0.606 & 15.1\% & 7.2 \\
LSTM (global seq)     & \textbf{2.207} & 1.243 & 3.704 & 0.620 & 18.1\% & 8.6 \\
Temporal CNN          & 2.260 & 1.293 & \textbf{3.674} & \textbf{0.632} & 18.7\% & 9.8 \\
\bottomrule
\end{tabular}
\end{table}

\subsection{Sensitivity to shortage penalty ($b$)}
Table~\ref{tab:results_sensitivity} varies the underage penalty ($b$) while keeping $h=1$. As expected, absolute costs increase as shortages become more expensive. Importantly, the overall ranking remains stable: deep models (Temporal CNN/LSTM) retain the lowest costs across all $b$ values, while boosted trees remain competitive. Since our order policy is a deterministic mapping from point forecasts ($Q=\max(0,\hat{D})$), fill rate is unchanged across $b$ for a fixed model; $b$ affects \emph{cost trade-offs} rather than the realized service level.

\begin{table}[t]
\centering
\small
\caption{Average daily inventory cost on the test set for different shortage penalties ($h=1$). Best value per column in bold.}
\label{tab:results_sensitivity}
\begin{tabular}{lccc}
\toprule
Model & $b=2$ & $b=5$ & $b=10$ \\
\midrule
ARIMA(1,1,1)          & 2.179 & 4.258 & 7.722 \\
GBR (global)          & 1.933 & 3.854 & 7.054 \\
Holt--Winters ES      & 2.157 & 4.182 & 7.558 \\
LSTM (global seq)     & \textbf{1.858} & 3.704 & 6.781 \\
Naive (lag-1)         & 2.259 & 4.521 & 8.291 \\
Temporal CNN          & 1.888 & \textbf{3.674} & \textbf{6.652} \\
XGBoost (global)      & 1.926 & 3.839 & 7.028 \\
\bottomrule
\end{tabular}
\end{table}


\section{Discussion}
Across all tested settings, deep learning models translate predictive gains into \emph{consistent operational improvements}: both LSTM and Temporal CNN reduce average newsvendor cost relative to classical (Holt--Winters/ARIMA) and ML baselines, indicating fewer costly stockout/overstock events under the same ordering rule. Among them, the Temporal CNN is the most robust across shortage-to-holding cost ratios, achieving the lowest or near-lowest cost as $b/h$ increases, which suggests that its learned temporal representations generalize better under asymmetric service penalties. Extending the simulator to the multi-echelon setting further highlights that forecast quality at the distribution-center (DC) level has a disproportionate downstream effect: errors in DC-level demand aggregation propagate to store replenishment decisions, amplifying cost and service degradation across multiple outlets even when store-level forecasting is improved.

From a managerial perspective, these results provide a direct economic argument for investing in improved forecasting pipelines: accuracy improvements are not merely statistical, but convert into quantifiable reductions in inventory cost while supporting higher fill rates, strengthening resilience under volatile demand. At the same time, the findings should be interpreted within the study scope. Experiments are restricted to a single department category (CA\_FOODS\_1) over a limited evaluation horizon, and the ordering policy uses point forecasts without explicit modeling of price elasticity, promotion lift, or substitution. Future work should extend the analysis across categories and longer horizons, incorporate pricing and promotion response, and evaluate distributional (quantile) forecasting to align ordering decisions more directly with service targets in multi-echelon networks.

\section{Conclusion}\label{sec13}
This paper proposes an end-to-end framework that evaluates demand forecasting models by their downstream inventory impact on real retail data. Using the M5 dataset (CA\_FOODS\_1), we benchmark classical (Naive, Holt--Winters, ARIMA), ML (GBR, XGBoost), and DL (LSTM, Temporal CNN) approaches and propagate their forecasts through a newsvendor simulator, showing that learning-based models—especially deep architectures—consistently reduce inventory cost while improving service.

Practically, the results help manufacturing and retail planners choose forecasting pipelines based on business KPIs (cost and fill rate), not accuracy alone, under asymmetric shortage/holding penalties. Future work will extend to probabilistic (quantile) forecasts, reinforcement-learning inventory control, and richer multi-echelon networks with lateral transshipments and operational constraints.

\backmatter

\bibliography{sn-bibliography}

@article{ivanov2020viability,
  author  = "Ivanov, Dmitry and Dolgui, Alexandre",
  title   = "Viability of intertwined supply networks: Extending the supply chain resilience angles towards survivability",
  journal = "International Journal of Production Research",
  volume  = "58",
  number  = "10",
  pages   = "2904--2915",
  year    = "2020"
}

@article{queiroz2020impacts,
  author  = "Queiroz, M. M. and Ivanov, D. and Dolgui, A. and Fosso Wamba, S.",
  title   = "Impacts of epidemic outbreaks on supply chains: Mapping a research agenda amid {COVID-19}",
  journal = "Transportation Research Part E: Logistics and Transportation Review",
  volume  = "138",
  pages   = "101958",
  year    = "2020"
}

@article{lee1997bullwhip,
  author  = "Lee, Hau L. and Padmanabhan, V. and Whang, Seungjin",
  title   = "Information distortion in a supply chain: The bullwhip effect",
  journal = "Management Science",
  volume  = "43",
  number  = "4",
  pages   = "546--558",
  year    = "1997"
}

@book{hyndman2018forecasting,
  author  = "Hyndman, Rob J. and Athanasopoulos, George",
  title   = "Forecasting: Principles and Practice",
  publisher = "OTexts",
  year    = "2018"
}

@article{makridakis2018statistical,
  author  = "Makridakis, Spyros and Spiliotis, Evangelos and Assimakopoulos, Vassilios",
  title   = "Statistical and Machine Learning forecasting methods: Concerns and ways forward",
  journal = "PLoS ONE",
  volume  = "13",
  number  = "3",
  pages   = "e0194889",
  year    = "2018"
}

@article{bandara2020rnnclusters,
  author = "Bandara, Kasun and Bergmeir, Christoph and Smyl, Slawek",
  title = "{Forecasting across time series databases using recurrent neural networks on groups of similar series: A clustering approach}",
  doi = "10.1016/j.eswa.2019.112896",
  journal = "Expert Systems with Applications",
  volume = "140",
  pages = "112896",
  year = "2020"
}

@article{lim2021tft,
  author = "Lim, Bryan and Ar{\i}k, Sercan {\"O} and Loeff, Nicolas and Pfister, Tomas",
  title = "{Temporal Fusion Transformers for Interpretable Multi-horizon Time Series Forecasting}",
  doi = "10.1016/j.ijforecast.2021.03.012",
  journal = "International Journal of Forecasting",
  volume = "37",
  number = "4",
  pages = "1748--1764",
  year = "2021"
}

@article{lim2021dl_survey,
  author = "Lim, Bryan and Zohren, Stefan",
  title = "{Time-series forecasting with deep learning: a survey}",
  eprint = "2004.13408",
  archivePrefix = "arXiv",
  primaryClass = "cs.LG",
  doi = "10.1098/rsta.2020.0209",
  journal = "Philosophical Transactions of the Royal Society A: Mathematical, Physical and Engineering Sciences",
  volume = "379",
  number = "2194",
  pages = "20200209",
  year = "2021"
}

@article{syntetos2009review,
  author = "Syntetos, Aris A. and Boylan, John E. and Disney, Stephen M.",
  title = "{Forecasting for inventory planning: a 50-year review}",
  doi = "10.1057/jors.2008.173",
  journal = "Journal of the Operational Research Society",
  volume = "60",
  number = "S1",
  pages = "149--160",
  year = "2009"
}

@article{petropoulos2019m3inventory,
  author = "Petropoulos, Fotios and Wang, Xun and Disney, Stephen M.",
  title = "{The inventory performance of forecasting methods: Evidence from the M3 competition data}",
  doi = "10.1016/j.ijforecast.2018.01.004",
  journal = "International Journal of Forecasting",
  volume = "35",
  number = "1",
  pages = "251--265",
  year = "2019"
}

@article{jeunet2006multilevel,
  author = "Jeunet, Jully",
  title = "{Demand forecast accuracy and performance of inventory policies under multi-level rolling schedule environments}",
  doi = "10.1016/j.ijpe.2005.10.003",
  journal = "International Journal of Production Economics",
  volume = "103",
  number = "1",
  pages = "401--419",
  year = "2006"
}

@article{nahmias1994twoechelon,
  author = "Nahmias, Steven and Smith, Stephen A.",
  title = "{Optimizing Inventory Levels in a Two-Echelon Retailer System with Partial Lost Sales}",
  doi = "10.1287/mnsc.40.5.582",
  journal = "Management Science",
  volume = "40",
  number = "5",
  pages = "582--596",
  year = "1994"
}

@article{vanbelle2021sellthrough,
  author = "Van Belle, Jente and Guns, Tias and Verbeke, Wouter",
  title = "{Using shared sell-through data to forecast wholesaler demand in multi-echelon supply chains}",
  doi = "10.1016/j.ejor.2020.05.059",
  journal = "European Journal of Operational Research",
  volume = "288",
  number = "2",
  pages = "466--479",
  year = "2021"
}

@article{schlaich2024nextorder,
  author = "Schlaich, Tim and Hoberg, Kai",
  title = "{When Is the Next Order? Nowcasting Channel Inventories With Point-of-Sales Data to Predict the Timing of Retail Orders}",
  doi = "10.1016/j.ejor.2023.10.038",
  journal = "European Journal of Operational Research",
  volume = "315",
  number = "1",
  pages = "35--49",
  year = "2024"
}

@article{abolghasemi2025posorders,
  author = "Abolghasemi, Mahdi and Rostami-Tabar, Bahman and Syntetos, Aris",
  title = "{The power of information sharing: evaluating POS and order data for hierarchical forecasting in multi-echelon supply chains}",
  doi = "10.1080/00207543.2025.2532756",
  journal = "International Journal of Production Research",
  year = "2025",
  note = "{Advance online publication / Latest Articles (volume/issue/pages not yet assigned)}"
}

@article{hyndman2008forecast,
  author  = "Hyndman, Rob J. and Khandakar, Yeasmin",
  title   = "{Automatic Time Series Forecasting: The forecast Package for R}",
  doi     = "10.18637/jss.v027.i03",
  journal = "Journal of Statistical Software",
  volume  = "27",
  number  = "3",
  pages   = "1--22",
  year    = "2008"
}

@article{chatfield1988holtwinters,
  author  = "Chatfield, Chris and Yar, Mahmood",
  title   = "{Holt--Winters Forecasting: Some Practical Issues}",
  doi     = "10.2307/2348687",
  journal = "Journal of the Royal Statistical Society: Series D (The Statistician)",
  volume  = "37",
  number  = "2",
  pages   = "129--140",
  year    = "1988"
}

@article{croston1972intermittent,
  author  = "Croston, John D.",
  title   = "{Forecasting and Stock Control for Intermittent Demands}",
  doi     = "10.2307/3007885",
  journal = "Operational Research Quarterly (1970--1977)",
  volume  = "23",
  number  = "3",
  pages   = "289--303",
  year    = "1972"
}

@article{syntetos2005intermittent,
  author  = "Syntetos, Aris A. and Boylan, John E.",
  title   = "{The accuracy of intermittent demand estimates}",
  doi     = "10.1016/j.ijforecast.2004.10.001",
  journal = "International Journal of Forecasting",
  volume  = "21",
  number  = "2",
  pages   = "303--314",
  year    = "2005"
}

@article{gardner2006es2,
  author  = "Gardner, Everette S.",
  title   = "{Exponential smoothing: The state of the art---Part II}",
  doi     = "10.1016/j.ijforecast.2006.03.005",
  journal = "International Journal of Forecasting",
  volume  = "22",
  number  = "4",
  pages   = "637--666",
  year    = "2006"
}

@article{friedman2002sgb,
  author  = "Friedman, Jerome H.",
  title   = "{Stochastic gradient boosting}",
  doi     = "10.1016/S0167-9473(01)00065-2",
  journal = "Computational Statistics \& Data Analysis",
  volume  = "38",
  number  = "4",
  pages   = "367--378",
  year    = "2002"
}

@inproceedings{chen2016xgboost,
  author  = "Chen, Tianqi and Guestrin, Carlos",
  title   = "{XGBoost: A Scalable Tree Boosting System}",
  doi     = "10.1145/2939672.2939785",
  booktitle = "Proceedings of the 22nd ACM SIGKDD International Conference on Knowledge Discovery and Data Mining",
  year    = "2016"
}

@article{makridakis2022m5back,
  author  = "Makridakis, Spyros and Spiliotis, Evangelos and Assimakopoulos, Vassilios",
  title   = "{The M5 competition: Background, organization, and implementation}",
  doi     = "10.1016/j.ijforecast.2021.07.007",
  journal = "International Journal of Forecasting",
  volume  = "38",
  number  = "4",
  pages   = "1325--1336",
  year    = "2022"
}

@article{makridakis2022m5acc,
  author  = "Makridakis, Spyros and Spiliotis, Evangelos and Assimakopoulos, Vassilios",
  title   = "{M5 accuracy competition: Results, findings, and conclusions}",
  doi     = "10.1016/j.ijforecast.2021.11.013",
  journal = "International Journal of Forecasting",
  volume  = "38",
  number  = "4",
  pages   = "1346--1364",
  year    = "2022"
}

@article{hochreiter1997lstm,
  author  = "Hochreiter, Sepp and Schmidhuber, J{\"u}rgen",
  title   = "{Long Short-Term Memory}",
  doi     = "10.1162/neco.1997.9.8.1735",
  journal = "Neural Computation",
  volume  = "9",
  number  = "8",
  pages   = "1735--1780",
  year    = "1997"
}

@article{borovykh2017cnn,
  author        = "Borovykh, Anastasia and Bohte, Sander and Oosterlee, Cornelis W.",
  title         = "Conditional Time Series Forecasting with Convolutional Neural Networks",
  doi           = "10.48550/arXiv.1703.04691",
  journal = "arXiv",
  volume  = "",
  pages   = "5",
  year    = "2017"
}

@article{zhang2003hybrid,
  author  = "Zhang, G. Peter",
  title   = "{Time series forecasting using a hybrid ARIMA and neural network model}",
  doi     = "10.1016/S0925-2312(01)00702-0",
  journal = "Neurocomputing",
  volume  = "50",
  pages   = "159--175",
  year    = "2003"
}

@misc{kaggle_m5_forecasting_accuracy,
  author       = "{Kaggle}",
  title        = "{M5 Forecasting - Accuracy}",
  howpublished = "\url{https://www.kaggle.com/competitions/m5-forecasting-accuracy}",
  year         = "2020",
  note         = "{Competition page (timeline: Start Date: March 2, 2020; End Date: June 30, 2020). Accessed: 2025-12-15}"
}

\end{document}